\documentclass[
]{ceurart}

\usepackage{listings}
\usepackage{xcolor}

\colorlet{punct}{red!60!black}
\definecolor{background}{HTML}{EEEEEE}
\definecolor{delim}{RGB}{20,105,176}
\colorlet{numb}{magenta!60!black}

\lstdefinelanguage{json}{
    basicstyle=\normalfont\ttfamily,
    numbers=left,
    numberstyle=\scriptsize,
    stepnumber=1,
    numbersep=8pt,
    showstringspaces=false,
    breaklines=true,
    frame=lines,
    backgroundcolor=\color{background},
    literate=
     *{0}{{{\color{numb}0}}}{1}
      {1}{{{\color{numb}1}}}{1}
      {2}{{{\color{numb}2}}}{1}
      {3}{{{\color{numb}3}}}{1}
      {4}{{{\color{numb}4}}}{1}
      {5}{{{\color{numb}5}}}{1}
      {6}{{{\color{numb}6}}}{1}
      {7}{{{\color{numb}7}}}{1}
      {8}{{{\color{numb}8}}}{1}
      {9}{{{\color{numb}9}}}{1}
      {:}{{{\color{punct}{:}}}}{1}
      {,}{{{\color{punct}{,}}}}{1}
      {\{}{{{\color{delim}{\{}}}}{1}
      {\}}{{{\color{delim}{\}}}}}{1}
      {[}{{{\color{delim}{[}}}}{1}
      {]}{{{\color{delim}{]}}}}{1},
}

\newcolumntype{P}[1]{>{\centering\arraybackslash}p{#1}}

\begin{document}

\copyrightyear{2024}
\copyrightclause{Copyright for this paper by its authors.
  Use permitted under Creative Commons License Attribution 4.0
  International (CC BY 4.0).}

\conference{Defactify 3: Third Workshop on Multimodal Fact Checking and Hate Speech Detection, co-located with AAAI 2024.}

\title{Overview of Factify5WQA: Fact Verification through 5W Question-Answering}

 \author[1]{Suryavardan Suresh}[email =  ss17323@nyu.edu]
 \address[1]{New York University, USA}

 \author[2]{Anku Rani}
 \address[2]{University of South Carolina, USA}
 
 \author[3]{Parth Patwa}
 \address[3]{UCLA, USA}
 
 \author[4]{Aishwarya Reganti}
 \address[4]{CMU, USA}

\author[5]{Vinija Jain\dag}
 \address[5]{Amazon AI, USA}

\author[4,5]{Aman Chadha\dag}
 \address[4]{Stanford University, USA}
 \address[5]{Amazon GenAI, USA}
 
 \author[2]{Amitava Das}[email =  amitava@mailbox.sc.edu]

 \author[2]{Amit Sheth}

 \author[6]{Asif Ekbal}
 \address[6]{IIT Patna, India}

\renewcommand{\thefootnote}{\fnsymbol{footnote}}
\footnotetext[2]{Work does not relate to the position at Amazon.}
\renewcommand*{\thefootnote}{\arabic{footnote}}
\setcounter{footnote}{0}

\begin{abstract}
Researchers have found that fake news spreads much times faster than real news \cite{doi:10.1126/science.aap9559}. This is a major problem, especially in today's world where social media is the key source of news for many among the younger population. Fact verification, thus, becomes an important task and many media sites contribute to the cause. Manual fact verification is a tedious task, given the volume of fake news online. The  Factify5WQA shared task aims to increase research towards automated fake news detection by providing a dataset with an aspect-based question answering based fact verification method. Each claim and its supporting document is associated with 5W questions that help compare the two information sources. The objective performance measure in the task is done by comparing answers using BLEU score to measure the accuracy of the answers, followed by an accuracy measure of the classification. The task had submissions using custom training setup and pre-trained language-models among others. The best performing team posted an accuracy of 69.56\%, which is a near 35\% improvement over the baseline.
\end{abstract}


\maketitle

\section{Introduction} 
Manual fact-checking is a laborious process where journalists must scour multiple online and offline sources, assess their reliability, and synthesize the information to reach a final verdict, often taking hours or days depending on the claim's complexity. With the rise of social media and rapid news dissemination, automated fact-checking has emerged as an important AI problem to combat the dangers of fraudulent claims masquerading as reality. As per surveys from Statista \cite{Statista}, no country had over of 80\% of its people trusting media, with the number being below 50\% in USA. 

The preceding paragraph highlights the importance of such tasks and the requirement for a capable automated fact verification pipeline. Aiming to encourage development of such pipelines, with the goal to have an automated model analogous to the manual process, the factify 1 \cite{factify,factify1overview} and factify 2 \cite{factify2,factify2overview} shared tasks were previously conducted. These tasks focused on multi-modal fact checking that relies on comparison i.e. an entailment based approach. Both tasks had image and text pairs for both a claim and a supporting document, where their relationship defined their label (Support Multimodal, Support Text, Insufficient Multimodal, Insufficient Text and Refute).

With the advent of Large Language Models (LLMs), we have seen highly capable language models. The generative abilities of such models are quite evident and widely used. Thus, it is apparent that LLMs or more generally generative models must be tested in the fact verification domain. The Factify5WQA shared task adds to the Factify task by presenting the 5W questions, such that the answers to these questions based on the claim and the ground truth answers we curated for the evidence document can be used for fact checking.

The paper is organized as follows: we describe the task details in section \ref{sec:task}. Section \ref{sec:related} mentions related work whereas  Section \ref{sec:systems} describes our baseline and the participants' system. The results are provided in section \ref{sec:systems} and finally we conclude in section \ref{sec:conclusion}.

\section{Related Work}
\label{sec:related}
Several datasets and shared tasks on fact verification have been introduced to benchmark advancements in automated fact-checking, encouraging the development of robust algorithms. Over the years, researchers have produced a wide range of datasets and articles addressing the many challenges involved in automated fact checking.

An avenue of research deals with the analysis of the claim without an associated evidence, some examples include analyzing linguistic characteristics, stylometry etc. \cite{wang2017liar, rashkin2017truth, stylometry}. There also exists active research towards multilingual claim detection \cite{stanceosaurus, hu-etal-2022-chef} and fact checking with respect to a specific domain \cite{mohr-etal-2022-covert, patwa2021fighting, patwa2021overview}. Multi-modal datasets have also been explored with datasets for image, audio and video based fact checking \cite{factify, liu2023covid, khalid2021fakeavceleb, nakamura2019Fakeddit}. Datasets with textual claim and supporting evidence to validate or refute the claim are predominantly used, including datasets that provide a synthetic claim for the evidence \cite{fever, feverous}. Shared tasks have also proven to be great avenues to introduce fact verification datasets and establish fact checking methodologies \cite{nakamura2019Fakeddit, fever, rumoureval, fnc1}. 

FAVIQ \cite{faviq} has claims authored by crowdworkers and the authors present a fact checking approach that uses information seeking questions to classify a given claim-evidence pair as fake or not. In Factify5WQA, we add to the fact checking task by incorporating 5W questions that help highlight relevant context, with respect to the claim. We integrate data from several benchmark fact-checking datasets and complement them with 5W questions and answers. Details of our dataset are provided in next section and in \cite{rani-etal-2023-factify}.
\section{Task Details}
\label{sec:task}
The Factify5WQA dataset \cite{rani-etal-2023-factify} was constructed with prior fact checking work as its backbone. The dataset was curated by manually inspecting and selecting a subset of claims from six existing fact-checking datasets - FEVER \cite{fever}, VITC \cite{vitc}, Factify 1.0, Factify 2.0, FaVIQ \cite{faviq}, and HoVer \cite{hover} - based on quality criteria like claim and evidence length, grammatical correctness, etc. Specifically, for FEVER and VITC, only claims from the train split were included. From Factify 1.0 and 2.0, the multimodal part was discarded, and only the text-based claims were used. For FaVIQ, the more challenging 'A' set of ambiguous questions was selected over the 'R' set of unambiguous question-answer pairs. The curation process ensured a high-quality dataset suitable for evidence-based, interpretable open-domain fact-checking.

Additionally, to mimic the real world distribution and to increase the variance within the textual data across these datasets, claims were paraphrased. Based on manual testing, some SOTA models were selected and alternate versions of the claims were generated. The next step in the task dataset preparation is the 5W questions and their respective answers. This was done through semantic role labeling through an off-the-shelf tool AllenNLP. This library helps identify important parts of an input text and assigns roles such that we can identify subsets of the text that are relevant to the 5W questions i.e. Who, What, When, Where and Why. More about the 5W question-answer pairs generation and other specifics provided in the data paper for Factify5WQA. Following is a brief description of the labels/classes defined in the dataset.

\textbf{Support}: The \underline{claim} and \underline{evidence} are about the same statement i.e. they describe a common event, person etc.\\
\textbf{Neutral}: The \underline{claim} and \underline{evidence} are about the similar but not the same statement i.e. they have common words but are not describing a common scenario.\\
\textbf{Refute}: The \underline{evidence} actively refutes or opposed the \underline{claim}, thus indicating that the claim is false. 

The data statistics are provided in \ref{tab:splits} Some examples from the dataset are provided below.

\begin{lstlisting}[language=json,firstnumber=1,caption={Examples from the Factify5WQA dataset for the Refute and Support category. The evidence refutes the claim in the first example as indicated by the contrasting answers to the first question. In the second example, both the claim and evidence talk about a London police officer being injured. Despite the "when" question having different answers, the task requires that we highlight the first two answers and classify it as Support.},captionpos=b]
[
    {
        "claim": "Andre Agassi won seven titles.",
        "evidence": "Andre Kirk Agassi born April 29 , 1970 -RRB- is an American retired professional tennis player and former World No. 1 who was one of the sport's most dominant players from the early 1990s to the mid-2000s . Generally considered by critics and fellow players to be one of the greatest tennis players of all time ...,
        "question": [ "How many titles did andre agassi win?", "Who won seven titles?" ],
        "claim_answer": [ "seven titles", "Andre Agassi" ],
        "evidence_answer": [ "eight-time Grand Slam champion", "Agassi" ],
        "label": "Refute"
    }
    ,
    {
        "claim": "London police officer seriously injured in machete attack during vehicle stop. https://t.co/tnCa0MK6R9",
        "evidence": "By Julia Hollingsworth, CNNUpdated 0758 GMT (1558 HKT) August 8, 2019 (CNN)A London police officer is in a critical condition after a driver he pulled over attacked him with a machete. ",
        "question": [ "How was a london police officer seriously injured?", "Who was seriously injured in a machete attack?", "When was the london police officer attacked?" ],
        "claim_answer": [ ": in machete attack", "London police officer", "during vehicle stop" ],
        "evidence_answer": [ "a driver he pulled over attacked him with a machete", "A London police officer", "August 8, 2019" ],
        "label": "Support"
    }
]
\end{lstlisting}

\begin{table}[h!]
    \centering
    \begin{tabular}{|c|c|c|}
    \toprule
    Split & Size \# & Category splits \\
    \toprule
    Train & 10500 & 3500,3500,3500 \\
    Val & 2250 & 750,750,750 \\
    Test & 2250 & 750,750,750 \\
    \bottomrule
    \end{tabular}
    \caption{The train-val-test splits of the dataset along with the division of the labels.}
    \label{tab:splits}
\end{table}

\subsection{Evaluation}
As described in the previous sub-section, the dataset contains a set of questions for each given sample along with answers based on the claim and evidence respectively. Further each sample is assigned a class with respect to the relation between the claim and evidence i.e. Support, Neutral or Refute. The approach we define to evaluate performance on this dataset is with the use of BLEU score. The average BLEU score for the answers from the claim and evidence are compared to a threshold. If it is crosses the threshold, which we set to 0.3, and the label prediction matches the test data, the prediction is considered correct. The final score for the task is simply the percentage of such predictions i.e. $\#\text{ of correct answers}\div \#\text{ total samples}$. 
\section{Participating systems}
\label{sec:systems}
\begin{figure}[b]
    \centering
    \includegraphics[width=0.8\columnwidth]{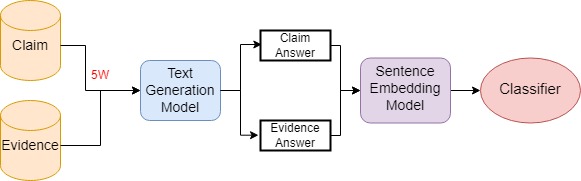}
    \caption{Pipeline for the baseline model for the Factify5WQA task}
    \label{fig:baseline}
\end{figure}

For the baseline model, we setup the pipeline shown in Figure \ref{fig:baseline}. We passed the claim and evidence to the Flan model \cite{flan} along with the 5W questions. For each question and claim/evidence pair, the prompt to the generative model is to generate an answer to the question based on context. The outputs from Flan are passed to the Mini-lm model in the Sentence Transformer library \cite{sbert} to generate embeddings for each answer. For the final predictions, we tried two approaches, i) Passing the Cosine similarity between claim and evidence answers to the classifier or ii) Passing the embeddings directly. Table \ref{tab:baselinescores} shows the results of our baseline pipeline experiments with different models used as classifiers. We can see that SVM classifier with cosine similarity gives the best results with a final score of 34.22\%.

With over 50 registrations in the competition web page, we had finals submissions from 3 teams with 2 of them making paper submissions.

The first of which is Team Trifecta \cite{trifecta}. They present “Pre-CoFactv3”, their custom architecture that uses ICL with a fine-tuned LLM for generation. They also introduce a model setup they call FakeNet - it leverages LLM's abilities along with co-attention for a final ensembled classification. Comprehensive experimental design and analysis demonstrating the effectiveness of the proposed methods, showcasing substantial improvements in accuracy over baselines. The results highlight the potential of the developed integration of LLMs and FakeNet for advancing open-domain fact verification.

The SRLFactQA \cite{srlfactqa}  team devised a Longformer-based SRL as input with Adapter-BERT used as the encoder. This was followed by attention based modules, which they refer to as the "Document Attention" module, to interpret the facts across the claim and evidence in-order to generate answers, before passing them to a classification module.

\begin{table}[]
\centering
\begin{tabular}{|l|r|}
\hline
\multicolumn{1}{|c|}{Classifier (input)} & \multicolumn{1}{c|}{Final score} \\ \hline
KNN (Embeddings)                         & 23.64\%                        \\ \hline
Logistic Regression (Embeddings)         & 24.53\%                        \\ \hline
Ridge (Embeddings)                       & 24.08\%                        \\ \hline
SVM (Embeddings)                         & 25.11\%                        \\ \hline
KNN (Cosine Sim)                         & 32.31\%                        \\ \hline
Logistic Regression (Cosine Sim)         & 31.95\%                        \\ \hline
Ridge (Cosine Sim)                       & 32.31\%                        \\ \hline
\textbf{SVM (Cosine Sim)}                & \textbf{34.22\%}               \\ \hline
\end{tabular}
\caption{Baseline scores for the pipeline shown in Figure \ref{fig:baseline}, with both the embeddings and the cosine scores between the embeddings used as inputs to the classifier.}
\label{tab:baselinescores}
\end{table}

\section{Results}
\label{sec:results}
\begin{table}[h]
    \centering
    \begin{tabular}{P{1cm}p{4cm}P{3cm}}
    \toprule
    Rank & Team & Final Score \\
    \toprule
    1 & \textbf{Team Trifecta \cite{trifecta}} & \textbf{69.56\%} \\
    2 & SRL\_Fact\_QA \cite{srlfactqa} & 45.51\% \\
    3 & Jiankang Han & 45.46\% \\
    4 & Baseline & 34.22\% \\
    \bottomrule
    \end{tabular}
    \caption{Leaderboard of the teams that made their final submissions to the Factify5WQA task.}
    \label{tab:task_results}
\end{table}

Table \ref{tab:task_results} shows the results all final submissions to the task along with the baseline. Team Trifecta \cite{trifecta} is the best performing team with an improvement of about 35\% over the baseline. They also outperform the team that places second in the shared task by over 20\%. The second and third team i.e. SRL\_Fact\_QA \cite{srlfactqa} and Jiankang Han, are seperated only by 0.05\%.

While all teams outperformed the baseline,  it can be seen in Table \ref{tab:taskindividual_results} that all participants had poor results for the Support category. On the other hand, all teams made the correct predictions on nearly 50\% of the Neutral or Refute samples, if not more. We note that, as per the BLEU scores, Team Trifecta got about 15\% of the generated answers incorrect while the other teams got 33\% incorrect. Finally, we can see that team trifecta has the best performance on all the classes. 

\begin{table}[h]
    \centering
    \begin{tabular}{P{1cm}p{4cm}P{2cm}P{2cm}P{2cm}}
    \toprule
    Rank & Team & Support & Neutral & Refute \\
    \toprule
    1 & Team Trifecta & \textbf{66.40\%} & \textbf{68.00\%} & \textbf{73.86\%} \\
    2 & SRL\_Fact\_QA & 36.13\% & 50.80\% & 49.60\% \\
    3 & Jiankang Han & 27.73\% & 59.20\% & 49.46\% \\
    4 & Baseline & 27.46\% & 32.93\% & 42.26\% \\
    \bottomrule
    \end{tabular}
    \caption{Leaderboard for each individual label with respect to the final submissions to the Factify5WQA task from Table \ref{tab:task_results}.}
    \label{tab:taskindividual_results}
\end{table}
\section{Conclusion and Future Work}
\label{sec:conclusion}

In this paper, we describe the the shared task Factify5WQA and provided a summary of participating systems. We saw that teams used LLMs or BERT. The best performing team achieved a score of 69.56\%, which shows that the problem remains unsolved. 

Future work could include expanding the 5wQA framework to multi-modality (text + images) and to other languages. 


\bibliography{main}

\end{document}